\pdfoutput=1

\documentclass[11pt]{article}

\usepackage{EMNLP2022}

\usepackage{times}
\usepackage{latexsym}
\usepackage{graphicx}
\graphicspath{ {images/} }
\usepackage{tabularx}

\usepackage[T1]{fontenc}

\usepackage[utf8]{inputenc}

\usepackage{microtype}

\usepackage{inconsolata}

\usepackage{multirow}
\usepackage{pifont}
\usepackage{enumitem}

\title{QUILL: Query Intent with Large Language Models using Retrieval Augmentation and Multi-stage Distillation}

\author{Krishna Srinivasan\thanks{\,\,\,\,Corresponding Authors} \\
  Google Research \\
  \texttt{krishnaps@google.com} \\\And
  Karthik Raman\footnotemark[1] \\
  Google Research \\
  \texttt{karthikraman@google.com} \\\And
  Anupam Samanta \\
  Google \\
  \texttt{anupamsamanta@google.com} \\\AND
  Lingrui Liao \\
  Google \\
  \texttt{lingrui@google.com} \\\And
  Luca Bertelli \\
  Google \\
  \texttt{lbertelli@google.com} \\\And
  Mike Bendersky \\
  Google Research \\
  \texttt{bemike@google.com} \\}

\begin{document}
\maketitle

\begin{abstract}

Large Language Models (LLMs) have shown impressive results on a variety of text understanding tasks. Search queries though pose a unique challenge, given their short-length and lack of nuance or context. Complicated feature engineering efforts do not always lead to downstream improvements as their performance benefits may be offset by increased complexity of knowledge distillation. Thus, in this paper we make the following contributions: (1) We demonstrate that Retrieval Augmentation of queries provides LLMs with valuable additional context enabling improved understanding. While Retrieval Augmentation typically increases latency of LMs (thus hurting distillation efficacy), (2) we provide a practical and effective way of distilling Retrieval Augmentation LLMs. Specifically, we use a novel two-stage distillation approach that allows us to carry over the gains of retrieval augmentation, without suffering the increased compute typically associated with it. (3) We demonstrate the benefits of the proposed approach (QUILL) on a billion-scale, real-world query understanding system resulting in huge gains. Via extensive experiments, including on public benchmarks, we believe this work offers a recipe for practical use of retrieval-augmented query understanding.

\end{abstract}

\section{Introduction}

The recent advent of billion+ parameter Large Language Models (LLMs) -- such as T5 \citep{t5-paper}, mT5 \citep{mt5-paper}, GPT-3 \citep{gpt3-paper} and most recently PaLM \citep{palm-paper} -- has disrupted many language understanding tasks -- with new benchmarks set or eclipsed routinely by these Transformer models and their variants.

Queries -- especially keyword search ones -- present a unique challenge though. Their short length, inherent ambiguity and lack of grammar mean query understanding tasks typically require more memorization and world knowledge than other NLP tasks~\citep{broder2007robust}. Consequently, despite LLMs leading performance on language and query understanding tasks -- like intent classification, query parsing and relevance prediction -- there is significant room for further improvement.


In this paper we leverage Retrieval-Augmentation to provide LLMs more context and grounding for search queries. We show that the titles and URLs of documents retrieved for the query, greatly help improve LLMs query understanding capabilities. While different retrieval augmentation models exist, we show that even simple concatenation of these titles / urls with the query can help improve LLM performance considerably.

However, the use of retrieval augmentation leads to a new challenge: Increased complexity of LLM inference. More specifically, the quadratic complexity of self-attention in Transformer models means that the latency of LLMs blows up given these (often 10x+) longer input sequences. This presents a significant problem as LLMs are impractical for online use and thus need to be distilled into smaller, more efficient models to be served online. However knowledge distillation \citep{gou2021knowledge} into these \emph{student} models requires a lot of distillation data annotated by these LLMs -- which may not be feasible for these retrieval augmented models.

Thus as a remedy we introduce a new two-stage distillation approach. In the first stage of this approach we distill the retrieval-augmented (long input) LLM (the \emph{Professor}) into a non-retrieval augmented (short input) LLM (the \emph{Teacher}) using a small distillation set. This second LLM \emph{Teacher} is in turn distilled into the final \emph{Student} using a large set.

Via extensive experiments on a large-scale, real-world problem and data we demonstrate that the resulting QUILL system provides for an efficient and effective way of retaining the performance gains of retrieval augmented LLMs on query understanding tasks.

\section{Related Work}

Large language models (LLMs) such as mT5~\cite{mt5-paper} demonstrated significant performance improvements on a variety of natural language understanding (NLU) tasks. Specifically in the context of query understanding, researchers found that (a) model size significantly effects the quality of the resulting models~\cite{Nogueira+al:2019,Han+al:2020}, and (b) using additional context in the form of query-associated documents is crucial to the model performance due to the paucity of context available in the query itself~\cite{nogueira2019doc2query,Zhang+al:2020}. Retrieval augmentation of the query with the search results retrieved by it is a proven way to incorporate such context in LLM training for NLU tasks, as has been shown recently by models such as RAG~\cite{lewis2020retrieval}, REALM~\cite{guu2020realm}, and RETRO~\cite{borgeaud2021improving}.

In this paper, we leverage this insight to improve performance of query intent prediction~\cite{broder2002taxonomy} --- a crucial query understanding task that is at the heart of modern search engines -- using LLMs. Prior work by ~\citet{broder2007robust} found importance of retrieval augmentation using statistical methods for this task. Statistical retrieval augmentation has also been found critical for other query understanding tasks including query expansion~\cite{broder+al:2008,diaz2006improving} and query tagging~\cite{Wang2020}. We demonstrate similar benefits when using retrieval-augmented LLMs as well.

We also leverage Knowledge Distillation ~\cite{hinton2015distilling,mirzadeh2020improved,gou2021knowledge} techniques to create a Student model that retains the LLMs gains.



\section{Query Intent Understanding}

While the techniques described in this paper could be applied to any query understanding task, for the sake of brevity we focus on the task of query intent classification. Query intent (QI) classification is a classical IR task studied for over two decades \cite{kang2003query, baeza2006intention, jansen2008determining, kathuria2010classifying, lewandowski2012deriving, figueroa2015exploring, mohasseb2019customised}. This task is particularly important in practice, as it is at the top of the search funnel, and the entire search engine behavior may vary based on the predicted query intent. Given the centrality of this task on overall retrieval, models for this task need to be both fast (\emph{i.e.,} low latency) and high efficacy. Thus even a single percentage point quality gain on the QI task can be considered a major accomplishment.

\begin{table}
\centering
\begin{tabular}{lc@{\hspace{0.75\tabcolsep}}c@{\hspace{0.75\tabcolsep}}c@{\hspace{0.75\tabcolsep}}c}
\hline
\textbf{Data} & \textbf{Train} & \textbf{Val} &
\textbf{Test} & \textbf{Unlabeled} \\
\hline

\textbf{Orcas-I} & 1.28M & 1K & 1K & 10.3M \\
\hline
\textbf{EComm} & 36K & 4K & 4K & 128M \\
\hline
\end{tabular}
\caption{Statistics of datasets used.}
\label{tab:dataset-stats}
\end{table}

In this paper we tackle the QI task using LLMs. In particular we use two datasets in our study whose details are provided in Table~\ref{tab:dataset-stats}:
\begin{itemize}
    \item \textbf{EComm}: Our main dataset will be a \textbf{real-world} dataset. Cast as a binary classification problem, this task involves identifying queries with a specific intent -- where the required intent is similar to the \emph{transactional} intent of the Broder taxonomy \cite{broder2002taxonomy} in the context of e-commerce. As common in real-world applications, the human labeled data is accompanied by a large unlabeled set -- that is used for knowledge distillation.
    \item \textbf{Orcas-I}: The largest publicly available query intent dataset is ORCAS-I \cite{Alexander_2022_orcas_i}. This comprises queries of the ORCAS dataset \cite{orcas_cikm_paper} labeled with one of 5 intent classes. Note that while the test set is human-labeled, the training set labels are weak labels as detailed in ORCAS-I \cite{Alexander_2022_orcas_i} paper's Methodology section.
\end{itemize}


\section{QUILL Methodology}
\label{sec:methodology}

\begin{figure*}[ht]
  \centering
  \includegraphics[width=0.9\textwidth]{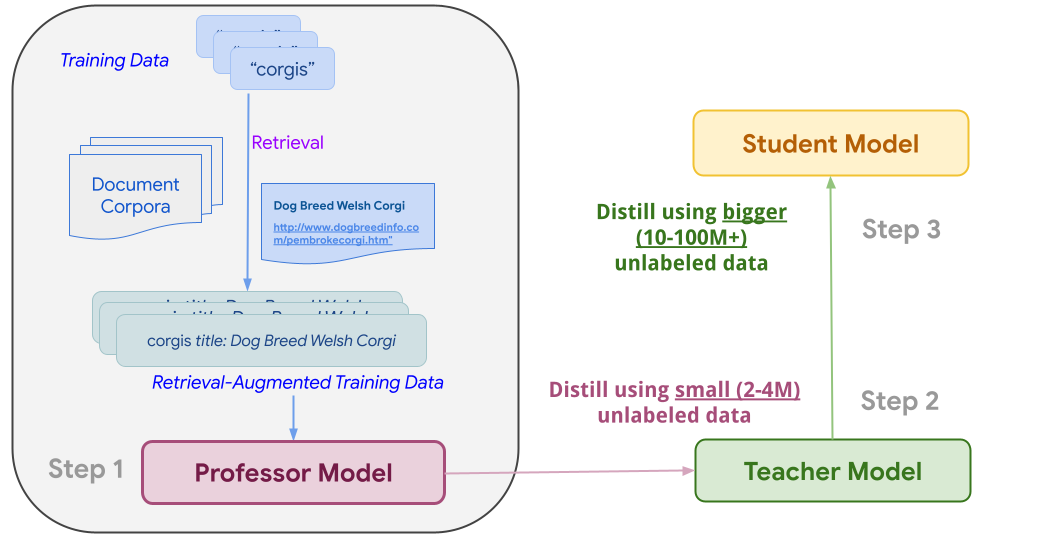}
  \caption{QUILL Architecture : Retrieval Augmentation and Multi-stage Distillation. }
  \label{fig:arch}
\end{figure*}

The keyword nature of queries and lack of context make the QI task (like other query understanding tasks) challenging for LLMs. Thus we propose QUILL as a solution. As seen in Figure \ref{fig:arch}, QUILL consists of two stages: (a) Retrieval Augmented LLM training, (b) Multi-Stage Distillation into efficient student.

\textbf{Retrieval Augmented (RA) LLM:} The key insight here is that titles / urls of related documents could provide valuable context to help understand the intent of the query. For example, it may not be immediately apparent what a query like \texttt{ua 1234} may mean. However, via the retrieved documents we can understand that the query is seeking information about a United Airlines flight. 

While there are multiple ways of augmenting the input via retrieved documents (example: the Fusion-in-Decoder architecture \cite{izacard2020_fid}), we chose to study the most straightforward and popular approach of concatenating the titles / urls of the retrieved top-k documents with the original query as the input to our LLM. As shown empirically (Sec~\ref{sec:experiments}), this model outperforms all baselines -- demonstrating the value of additional context.

\begin{small}
\begin{table}
\centering
\begin{tabular}{l @{\hspace{0.75\tabcolsep}}|c@{\hspace{0.5\tabcolsep}}c|c@{\hspace{0.5\tabcolsep}}c}
\hline
\textbf{Feature} & \multicolumn{2}{c|}{\textbf{EComm}} & \multicolumn{2}{c}{\textbf{Orcas-I}}
\\
 & Median & 99\% & Median & 99\% \\
\hline
Query & 5 & 18 & 5 & 10 \\
ExpandTerms & 17 & 28 & N/A & N/A \\
(Up to) 10 Titles & 157 & 245 & 13 & 104 \\
(Up to) 10 URLs & 159 & 304 & 39 & 238 \\
\hline
\end{tabular}
\caption{mT5 sequence lengths by features.}
\label{tab:seq-len-stats}
\end{table}
\end{small}

\textbf{Multi-stage distillation:} The drawback of RA is the additional sequence length of the input. As seen from Table~\ref{tab:seq-len-stats}, augmenting a query with (upto) 10 titles and urls increases the sequence length by an order of magnitude. Consequently, this makes distillation far more challenging given the quadratic complexity of sequence length (due to self-attention) in transformer models. This leaves us in a dichotomy between a more effective model with a much smaller distillation set, vs. a lower performing model with a larger distillation set. Given a large distillation set is required for training an effective student, this leaves us at risk of not being able to benefit from RA, given that a very large dataset with RA will incur very long and impractical inference times.

To get the best of both worlds we propose a two-stage distillation approach. In the first stage we distill the \emph{Professor} RA LLM into a \emph{Teacher} LLM without RA. The Teacher model uses ExpandTerms which provide additional context to the queries. While this may not be as expressive as retrieval augmentation, this provides a good compromise of greatly reducing sequence length while giving up only a little in performance. We do so by using a small subset of the unlabeled data. As shown empirically, a LLM teacher trained in this manner performs significantly better than a non-RA LLM trained directly on the human data, while at the same time allowing us to efficient distillation.

In the second stage we use the \emph{Teacher} LLM to annotate the entire unlabeled dataset. This is in turn used to train the final \emph{Student} model that will be used in practice. 

\section{Experiments and Results}
\label{sec:experiments}

\textbf{Experimental Setup:} Our experiments were all conducted using the mT5 \cite{mt5-paper} checkpoints. We validate performance across three learning rates (1e-3, 1e-4, 5e-5) -- selecting the best checkpoint using the validation set loss. For models trained from the provided training sets, we used a batch size of 64 in our experiments and  trained for 4K steps (EComm) / 20K steps (Orcas-I). 

For distilled models, we used a batch size of 128 for Teacher models and 1024 for Student models. We use different batch sizes because of the model architectures, mT5 for the Teacher vs a BERT-based model for the Student. In both cases, we  trained for 1 epoch, unless mentioned otherwise. We only use the encoder of the mT5 model with an additional layer added on top to predict the classification scores. The Professor, the Teacher and the Student fine-tuning experiments are all set up as a query intent classification task. Given that the Teacher and Student models are trained on millions of examples and this itself is a time and resource intensive step, we restrict our experiments to only one epoch. We demonstrate performance gains even with one epoch via the techniques elaborated in this paper.

We studied the effect of distillation data size, for both stages of distillation. For the EComm dataset, we used an in-house retriever to find related documents. For Orcas-I, we use the provided docids (aggregated at per-query level) for retrieval augmentation. Unless specified, we use (upto) the top-10 results for retrieval augmentation\footnote{For Orcas-I, nearly 2/3rd of the queries only have a single provided result, while some have upwards of 2000 results, which is why the lengths for RA features on Orcas-I in Table~\ref{tab:seq-len-stats} are smaller. The 10 results augmented are randomly chosen if more exist.}. Sequence length for models are based on the training set and features (set to 99\%-percentile of sequence lengths).


\textbf{Students and Features:}
Our experiments demonstrate results for a fast, efficient 4-layer transformer student architecture, with hidden dimensionality of 256.
We default to using the query as the only feature in the student for simplicity. To compare against query expansion techniques, we used a sophisticated in-house memorization-based query expansion model in our Professor / Teacher experiments on EComm. This expansion model -- which we refer to as \textbf{ExpandTerms} -- provides a list of related terms for a given query, which are concatenated with the query (and identifiers for start / end of each feature).

\textbf{Metrics:}
To compare performance of different models we use two metrics: \textbf{MicroF1} and \textbf{MacroF1} for Orcas-I, and \textbf{AUC-PR} and \textbf{AUC-ROC} for EComm. For EComm, we only report performance of models relative to the mT5 query-only Base-sized model\footnote{For a sense of scale, each 0.5\% point increase in metrics on EComm is considered a significant gain.}.

\begin{table}[t]
\centering
\begin{tabular}{l@{\hspace{0.75\tabcolsep}}c@{\hspace{0.75\tabcolsep}}|c@{\hspace{0.75\tabcolsep}}c@{\hspace{0.75\tabcolsep}}}
\textbf{Model} & \textbf{Size} & \textbf{ROC} & \textbf{PR}\\
\hline
query & Base & 0.0\% & 0.0\% \\
\ + RA (titles, urls) & Base & \textbf{\textcolor{teal}{+4.3\%}} & \textbf{\textcolor{teal}{+4.6\%}} \\
\hline
query & XL & \textcolor{teal}{+2.7\%} & \textcolor{teal}{+3.1\%} \\
\ + RA (titles, urls) & XL & \textbf{\textcolor{teal}{+6.3\%}} & \textbf{\textcolor{teal}{+6.7\%}} \\
\hline
query & XXL & \textcolor{teal}{+3.0\%} & \textcolor{teal}{+3.3\%} \\
\ + RA (titles, urls) & XXL & \textbf{\textcolor{teal}{+6.4\%}} & \textbf{\textcolor{teal}{+6.9\%}} \\
\hline

\hline

\end{tabular}
\caption{\label{exp-results-professor1-ecomm}
Results demonstrating the benefit of Retrieval Augmentation (RA) across all model sizes.
}
\end{table}

\subsection{Effect of Retrieval Augmentation}
While the use of retrieval augmentation (RA) has been known to improve query classification performance \cite{broder2007robust}, the benefit of RA is unclear in the age of LLMs. 
Thus, we start by evaluating the first stage of QUILL \emph{i.e., the RA model}.
As seen in Table~\ref{exp-results-professor1-ecomm}, RA improves performance significantly across all model sizes including the billion-parameter+ XL and XXL models. In fact the gains from RA on the Base-sized model exceed the gains obtained by increasing model size of a query-only model to XXL.
Given the gains observed across all models sizes, we use Base-sized models in the rest of the paper to simplify experimentation.

\begin{table}
\centering
\begin{tabular}{l@{\hspace{0.75\tabcolsep}}|c@{\hspace{0.75\tabcolsep}}c}
\textbf{EComm} & \textbf{ROC} & \textbf{PR}\\
\hline
query & 0.0\% & 0.0\% \\
\ + Terms  & \textcolor{teal}{+2.6\%} & \textcolor{teal}{+1.9\%} \\
\ + RA (titles)  & \textcolor{teal}{+4.8\%} & \textcolor{teal}{+4.8\%} \\
\ + RA (titles) + Terms  & \textcolor{teal}{+5.1\%} & \textcolor{teal}{+5.2\%} \\
\ \textbf{+ RA (urls)} & \textbf{\textcolor{teal}{+5.3\%}} & \textbf{\textcolor{teal}{+5.7\%}} \\
\hline
\end{tabular}
\caption{\label{exp-results-professor1-features}
Analysis of the impact of different features (using Base-sized models) for the EComm dataset. ExpandTerms abbreviated as Terms.
}
\end{table}

\begin{table}
\centering
\begin{tabular}{l@{\hspace{0.75\tabcolsep}}|c@{\hspace{0.75\tabcolsep}}c}
\textbf{EComm} & \textbf{ROC} & \textbf{PR}\\
\hline
\textbf{Orcas-I} & \textbf{MicF1} & \textbf{MacF1}\\
\hline
query & 69.8 & 69.75 \\
\ + RA (titles)  & \textcolor{teal}{+6.3\%} & \textcolor{teal}{+5.1\%} \\
\ \textbf{+ RA (urls)}  & \textbf{\textcolor{teal}{+8.2\%}} & \textbf{\textcolor{teal}{+6.2\%}} \\
\ + RA (titles+urls) & \textbf{\textcolor{teal}{+9.0\%}} & \textbf{\textcolor{teal}{+7.2\%}} \\
\hline

\end{tabular}
\caption{\label{exp-results-professor1-features-orcas}
Analysis of the impact of different features (using Base-sized models) for the Orcas-I dataset.
}
\end{table}

A natural question that may arise though is how do these gains from RA compare to those obtained by powerful query expansion techniques. Thus, we performed an in-depth ablation of features for the EComm dataset (on a Base-sized model for ease of experimentation) as seen in Table~\ref{exp-results-professor1-features} and for the Orcas-I dataset as seen in Table~\ref{exp-results-professor1-features-orcas}. These results clearly demonstrate the potency of powerful query expansion models (\emph{i.e.,} \textbf{ExpandTerms}) -- as evidenced by the large $\sim$2\% gains over query-only models. However, we find that RA adds even more value over these highly sophisticated expansion models with an a nearly 5+\% increase in performance. Furthermore, we find that RA techniques can still be combined with query expansion for further gains.

The improvements on RA for Orcas-I (seen in Table~\ref{exp-results-professor1-features-orcas}) are even more substantial, with a nearly 9\% improvement over the query-only baseline, via the use of the titles and urls of related documents. Interestingly, among RA features we find that urls tend to perform slightly better than titles on both datasets. We believe this to be because titles can have a higher variance of informativeness -- with both highly verbose and very short titles commonly seen. Hence, given the simplicity and consistency of urls, we chose to use RA(urls) for subsequent experiments as the \emph{Professor} model.


\begin{table}
\centering
\begin{tabular}{c@{\hspace{0.25\tabcolsep}}|c@{\hspace{0.75\tabcolsep}}c}
\textbf{EComm} & \textbf{ROC} & \textbf{PR}\\
\hline
\textbf{Baseline Teacher}  & \multirow{2}{*}{{+2.6\%}} & \multirow{2}{*}{{+1.9\%}} \\
(Finetuned on Training Set) & & \\
\hline

QUILL Teacher  & \multirow{2}{*}{\textcolor{teal}{+3.3\%}} & \multirow{2}{*}{\textcolor{teal}{+2.8\%}} \\
(2M Prof Distilled Set) & & \\
\textbf{QUILL Teacher (4M)} & \textbf{\textcolor{teal}{+3.4\%}} & \textbf{\textcolor{teal}{+2.9\%}} \\
QUILL Teacher (8M)& \textcolor{teal}{+3.5\%} & \textcolor{teal}{+2.9\%} \\
\hline
\\
\textbf{{\emph{QUILL Professor}}}  & \textbf{\textcolor{teal}{+5.3\%}} & \textbf{\textcolor{teal}{+5.7\%}} \\
\\
\end{tabular}
\caption{\label{tab:exp-results-teacher-training-ecomm}
Comparison of different Teacher models trained directly or via Professor-distillation for the EComm dataset.
}
\end{table}

\begin{table}
\centering
\begin{tabular}{c@{\hspace{0.25\tabcolsep}}|c@{\hspace{0.75\tabcolsep}}c}
\textbf{Orcas-I} & \textbf{MicF1} & \textbf{MacF1} \\ \hline
\textbf{Baseline Teacher}  & \multirow{2}{*}{69.8} & \multirow{2}{*}{69.75} \\
(Finetuned on Training Set) & & \\ \hline
QUILL Teacher  & \multirow{2}{*}{\textbf{\textcolor{teal}{+1.1\%}}} & \multirow{2}{*}{\textbf{\textcolor{teal}{+0.8\%}}} \\
(2M Prof Distilled Set) & & \\
\hline

\end{tabular}
\caption{\label{tab:exp-results-teacher-training-ecomm-orcas}
Comparison of different Teacher models trained directly or via Professor-distillation for the Orcas-I dataset.
}
\end{table}


\subsection{Distilling gains from RA}

We next focus on the second stage of QUILL: Distilling the RA model. Typically larger amounts of distillation data lead to better performance. However, given the increased sequence length of RA models \underline{and} the cost of retrieval augmentation itself, annotating large distillation sets is highly challenging. Thus to capture such practical trade-offs, we only used a small subset of the unlabeled data for the QUILL \emph{Professor} to \emph{Teacher} distillation. In particular, we used 4M examples for EComm (\emph{i.e.,} 3.1\% of unlabeled data) and 2M for Orcas-I (19\%) for this first stage of distillation -- to represent a set that is small enough set to be practical,  but large enough to learn from. However, we do share results for varying this size to understand its importance.

QUILL \emph{Teacher} models were thus trained by distilling the RA(urls) \emph{Professor} models. Our \emph{Teacher} models had the same capacity and architecture (\emph{i.e., mT5-Base} as the Professor\footnote{We observed similar trends even if the Teacher had less capacity than the Professor.} -- except it does not use RA (features). 

As a realistic and competitive baseline, we chose a \emph{Baseline Teacher} that resembles the QUILL \emph{Teacher} in all aspects bar one -- the data they are trained on. Specifically, the Baseline Teacher is directly trained from the gold-labeled training data, unlike the QUILL teacher. We believe this is representative of practical applications today, where LLMs are trained directly on gold-labeled sets (before being distilled into the final student models). To further challenge QUILL, we leverage the powerful \textbf{ExpandTerms} features (for the EComm dataset) in our Teacher models -- both Baseline and QUILL. We believe this provides a more challenging but realistic evaluation setup, since many baseline models in use today avail of powerful features (along with the query).

As seen from the results in Table~\ref{tab:exp-results-teacher-training-ecomm} and Table~\ref{tab:exp-results-teacher-training-ecomm-orcas}, we find the QUILL Teachers provide a significant performance improvement over the Baseline Teacher, despite having never directly seen the gold label data. On EComm, despite using an enhanced (realistic) baseline, QUILL teachers are $\sim$1\% better on all metrics. We find a similar gap on Orcas-I despite the Teacher there being trained on only 2M examples (just 1.5x the training set size). Put differently, we now have trained our non-retrieval augmented language model to benefit from the gains of retrieval augmentation. Even though the student model does not have Retrieval Augmentation, because of the Teacher model’s performance improvement, it is possible to annotate a considerably large number of training examples. We observe the Student models to close the gap (compared to the Teacher) given larger training datasets.

To test the robustness of QUILL teachers we also varied the amount of distillation data used -- halving or doubling it. While there still exist distillation gaps to the professor (which can be narrowed via more distillation data) on both datasets, our proposed approach works well even when using small amounts of distillation data -- which in turn allows us to save significant compute.

\begin{table}
\centering
\begin{tabular}{l@{\hspace{0.25\tabcolsep}}|c@{\hspace{0.25\tabcolsep}}c@{\hspace{0.25\tabcolsep}}}
\textbf{Model (\# Distillation)} & \textbf{ROC} & \textbf{PR}\\
\hline
No Distillation Student  & \textcolor{magenta}{-6.3\%} & \textcolor{magenta}{-7.3\%} \\
\hline
Baseline Student  & \textcolor{magenta}{-0.9\%} & \textcolor{magenta}{-1.6\%} \\
\hline
\hline
QUILL Student  & \textbf{\textcolor{teal}{+2.0\%}} & \textbf{\textcolor{teal}{+1.5\%}} \\
\hline
QUILL 1-Stage Student(4M)  & \textcolor{teal}{+0.4\%} & \textcolor{magenta}{-0.2\%} \\
QUILL 1-Stage Student(32M)  & \textcolor{teal}{+1.1\%} & \textcolor{teal}{+0.6\%} \\
\hline


\end{tabular}
\caption{\label{tab:exp-results-student-training}
Performance of the different student models trained from different teachers and using differing amounts of distillation data (on EComm).
}
\end{table}

\begin{table}[ht]
\centering
\begin{tabular}{p{0.12\textwidth}@{\hspace{0.15\tabcolsep}}|p{0.24\textwidth}@{\hspace{0.15\tabcolsep}}|c@{\hspace{0.15\tabcolsep}}}
\textbf{Query} & \textbf{Example URL} & W/L
\\
\hline
bengals & sports.yahoo.com/ nfl/teams/cin/ & \ding{51} \\
\hline
pah compounds & en.wikipedia.org/ wiki/Polycyclic\_aro matic\_hydrocarbon & \ding{51} \\
\hline
launch tech usa & launchtechusa.com/ & \ding{51} \\
\hline
\hline
noun university & en.wikipedia.org/ wiki/Noun & \ding{55} \\
\hline
airbed uk & www.airbnb.co.uk/  & \ding{55} \\
\hline
\end{tabular}
\caption{\label{tab:wins-losses-examples}
Wins/losses examples on Orcas-I.
}
\end{table}

\subsection{Final student training}

So far, we have shown that QUILL can learn a better (non-RA) teacher. However, an important question remains unanswered: Can these Teacher gains  be translated to the final student model? In particular, we postulate that the predictions of the QUILL Teacher may be more robust and easier to learn (for student models) than those of the Baseline. To verify this hypothesis we compared 4 fast student models (4-layer encoder-only models), with the only difference being the data they were trained on:
\begin{itemize}
    \item \emph{No Distillation Student}: This is the simple solution of directly training the Student using the labeled data.
    \item \emph{Baseline Student}: This is the current standard involving distilling the Baseline Teacher model using the full unlabeled set.
    \item \emph{QUILL Student}: This is the proposed solution involving distilling the QUILL Teacher model using the full unlabeled set.
    \item \emph{QUILL 1-Stage Student}: Rather than the two stage distillation approach, this student is directly distilled from the Professor using a  subset of the unlabeled data.
\end{itemize}

As seen from Table~\ref{tab:exp-results-student-training}, all QUILL-based students significantly outperform the Baseline Student. In particular our proposed 2-stage approach leads to a $\sim$3 point gain on both metrics. This is notable in that the gap between QUILL and Baseline students is even higher than the Teachers -- which we attribute to the QUILL Teacher labels being more robust.

Comparing different QUILL students, we find that there is a notable performance gain by first distilling into a non-RA teacher, before distilling into the final student. While 1-stage distillation performance improves as more data is used, even when 1/4th of all unlabeled data is retrieval-augmented and annotated by the Professor for direct distillation, it still falls short of the 2-stage approach. Together, these results show: (1) QUILL students outperform the current state-of-the-art significantly, and (2) QUILL benefits from the 2 stage distillation of Professor to Teacher to final student.

\subsection{Examples of Wins/Losses from RA }

While the previous sections focused on demonstrating the efficacy (and efficiency) of QUILL, we wanted to also understand \textbf{why} and \textbf{where} are some of these gains from RA stem from. To do so we used the test-set of Orcas-I and sampled illustrative examples of wins / losses (Table~\ref{tab:wins-losses-examples}) between the baseline and the retrieval-augmented professor models. One common win pattern we found for RA models is when the query is unclear, or uses technical terms / abbreviations. In these cases, the augmented urls / titles help provide additional context for the language model to understand what the query is about. On the flip side, we also found the biggest loss pattern to be when retrieval was inaccurate, which in turn misled the model regarding the query intent. For example, we found our retriever returned wikipedia results more often than it should, which mislead the model to believe the query had \emph{Factual} intent.

\section{Future Work}

While we studied the problem of query intent classification in this paper, the approach proposed in our paper is general and could be applied to any query understanding task. Following our approach, could enable myriad query understanding tasks use retrieval augmentation in a practically realistic and efficient manner. We leave this to future work though. We should also note that our experiments reveal non-trivial distillation gaps in both stages of distillation, which we believe is another open opportunity for future improvements.

\section{Conclusion}

This paper provides a practical recipe for combining Retrieval Augmentation and Large Language Models. In particular, we proposed QUILL as an approach to tackle the problem of query intent classification. Our empirical study demonstrates conclusively that Retrieval Augmentation can provide significant value over existing approaches. Furthermore we show that via our two-stage distillation approach, that QUILL not only learns better performing, more robust teachers, but also leads to even bigger gains when distilled into fast, real-world capable production student models.

\section{Acknowledgements}

We sincerely thank Jiecao Chen, William Dennis Kunz, Austin Tarango, Lee Gardner, Yang Zhang, Constance Wang, Derya Ozkan, Nitin Nalin, Raphael Hoffmann, Iftekhar Naim, Siddhartha Brahma, Siamak Shakeri, Hongkun Yu, John Nham, Ming-Wei Chang, Marc Najork, Corinna Cortes and many others for their insightful feedback and help. We also thank the EMNLP Reviewers for their thorough review, feedback and suggestions.

\newpage
\clearpage

\section*{Limitations}

This paper focuses on efficient and effective way of improving query intent classification using Retrieval Augmentation (RA) and Multi-stage distillation. While we have made the best attempts to ensure a robust and efficient method, we would be remiss to not point out some key limitations of our work:
\begin{itemize}
    \item \textbf{Quality of Retrieval:} A key reason for the gains seen in this paper is the use of Retrieval Augmentation. This additional context provided in the form of result titles / URLs are helpful, but are dependent on  the quality of the retrieval system. While we did not get a chance to explore the dependence of performance gains on retrieval quality, we plan to explore this in future work.
    \item \textbf{Dependency of Retrieval:} While our approach provides for a practical and low-compute way of incorporating retrieval augmentation, it still does add some compute (to augment the datasets) and system complexity. While we considered this trade-off well worth it in our use case, this may depend on specific settings.
    \item \textbf{Retrieval-Augmentation techniques:} As discussed in Section~\ref{sec:methodology}, we used a simple concatenation based retrieval augmentation. However, there do exist more sophisticated techniques for retrieval augmentation. For example, models built on a Fusion-in-Decoder \cite{izacard2020_fid} backbone have demonstrated great performance \cite{hofstatter2022multi,izacard2022few} and improved efficiency \cite{hofstatter2022fid}. We believe that these more sophisticated retrieval-augmentation technique may bring further improvements in our system and leave this for future work to follow up on.
    \item \textbf{Datasets:} The lack of large public query sets means that we were very limited in terms of what public benchmarks we could study this problem on. While ORCAS-I is the largest such available set, they lack many alternatives that are large enough to study the effects of distillation. In the future though, we hope to use the (somewhat related) problem of question-answering where larger datasets (with large enough unlabeled data) exist for a more thorough study.
    \item \textbf{Distillation gaps:} Our results also clearly demonstrate large distillation gaps in both stages. While there have been innovative techniques proposed to improve distillation performance, we intentionally chose to keep things simple as those approaches are largely complementary to the problem we study in this work.
    \item \textbf{Limited "Large" Model Experiments:} While our work is intended for and positioned in the context of "Large" Language Models, we realize that our most common model choice (mT5-Base), may not be the most representative model in that category. This was an intentional choice on our end as we hoped doing so would make the work more relevant to use cases and applications with more limited compute. For practitioners interested in models with tens of billions of parameters, we refer them to our analysis of mT5-XXL sized models in Table 3, that demonstrates the viability of our approach on models of that scale. 
    
\end{itemize}




\section*{Ethics Statement}

In this paper, we used only publicly available Language Model and Checkpoints that have been previously published -- namely mT5.

An important consideration when working with query datasets is data privacy. This is perhaps the biggest reason why there do not exist many large public query datasets. We intentionally chose ORCAS-I for this reason, as it is constructed from the ORCAS query set -- which is widely regarded as a well-constructed, non-PII, sufficiently anonymized query dataset. While the EComm dataset used in this paper is proprietary, we should note that it too has been scrubbed of PII and aims to follow the same (if not higher) data privacy principles. Our data (and methodology) do not contain any information for or target any demographic or identity characteristics.


The task we focus on -- query classification -- is a general problem that benefits everyone. In fact, it can enable better IR systems thereby benefiting users who otherwise might not get answers. Thus, we do not anticipate any biases or misuse issues stemming from this. We believe that by using publicly available and vetted retrieval models, the resulting retrieval augmented models should not create any new or further any existing biases.


In many ways a goal of our work is making retrieval augmentation more practical and reducing compute needs for any such applications. While we did present results with XXL sized models, we focused most of our experiments on the smaller, more efficient Base-sized models so as to benefit a wider section of our community and to reduce the computational needs of our experiments.


\bibliography{quill}
\bibliographystyle{acl_natbib}


\end{document}